\documentclass[letterpaper]{article} % DO NOT CHANGE THIS
\usepackage{aaai23}

\usepackage{times}  % DO NOT CHANGE THIS
\usepackage{helvet}  % DO NOT CHANGE THIS
\usepackage{courier}  % DO NOT CHANGE THIS
\usepackage[hyphens]{url}  % DO NOT CHANGE THIS
\usepackage{graphicx} % DO NOT CHANGE THIS
\usepackage{natbib}  % DO NOT CHANGE THIS AND DO NOT ADD ANY OPTIONS TO IT
\usepackage{caption} % DO NOT CHANGE THIS AND DO NOT ADD ANY OPTIONS TO IT
\usepackage{hyperref}
\usepackage{url}
\newcommand{\be}{\begin{equation}}
\newcommand{\ba}{\begin{array}{c}}
\newcommand{\ee}{\end{equation}}
\newcommand{\ea}{\end{array}}
\usepackage{amsmath,bm}
\usepackage{amsfonts}
\usepackage{graphicx,subfigure}
\usepackage{amsfonts,amsmath,amssymb, mathrsfs, amsthm}
\usepackage{multirow}
\usepackage{epic,eepic,eepicemu}
\usepackage{epsf}
\usepackage{epsfig}
\usepackage{graphics}
\usepackage{mathrsfs}
\usepackage{booktabs}
\usepackage{wrapfig}
\usepackage{psfrag}
\usepackage{bbold}
\usepackage{url}
\usepackage{tikz}
 
\usepackage[ruled,vlined]{algorithm2e}
\usepackage{xcolor}

\DeclareMathOperator*{\argmin}{arg\,min}

\usetikzlibrary{arrows}
\usepackage{verbatim}

\usepackage{bm}

\newcommand{\Gc}{\mathcal{G}}

\newcommand{\Nc}{\mathcal{N}}
\newcommand{\Lc}{\mathcal{L}}
\newcommand{\Xc}{\mathcal{X}}

\newcommand{\xb}{\bm{x}}

\newcommand*{\rom}[1]{\expandafter\@slowromancap\romannumeral #1@}
\ifodd 1 \newcommand{\rev}[1]{{\color{black}#1}}
\else \newcommand{\rev}[1]{#1} \fi

\ifodd 1 
\else  \fi

\SetKwInput{KwInput}{Input}                % Set the Input
\SetKwInput{KwOutput}{Output}

\title{Symbolic Metamodels for Interpreting Black-boxes Using Primitive Functions}

\author{
    Mahed Abroshan, \textsuperscript{\rm 1}
    Saumitra Mishra,\textsuperscript{\rm 2,}\thanks{Work done as a research associate at the Alan Turing Institute}
    Mohammad Mahdi Khalili  \textsuperscript{\rm 3,4}
}
\affiliations{
    \textsuperscript{\rm 1} The Alan Turing Institute, London, UK\\
    \textsuperscript{\rm 2} JP Morgan AI Research, London, UK\\
    \textsuperscript{\rm 3} Yahoo! Research, NYC, NY, USA\\
    \textsuperscript{\rm 4} CSE Department, The Ohio State University, Columbus, Ohio, USA\\
    mabroshan@tuirng.ac.uk, saumitra.mishra@jpmorgan.com, mahdi.khalili@yahooinc.com.
}

\begin{document}

\maketitle

\begin{abstract}
One approach for interpreting black-box machine learning models is to find a global approximation of the model using simple interpretable functions, which is called a metamodel (a model of the model). Approximating the black-box with
a metamodel can be used to 1) estimate instance-wise feature importance; 2) understand the functional form of the model; 3) analyze feature interactions. In this work, we propose a new method for finding interpretable metamodels. Our approach utilizes Kolmogorov superposition theorem, which expresses multivariate functions as a composition of univariate functions (our primitive parameterized
functions). This composition can be represented in the form of a tree. Inspired by symbolic regression, we use a modified form of genetic programming to search over different tree configurations. Gradient descent (GD) is used to optimize the parameters of a given configuration. Our method is a novel memetic algorithm that uses GD  not only for training numerical constants but also for the training
of building blocks. Using several experiments, we show that our method outperforms recent metamodeling approaches suggested for interpreting black-boxes.
\end{abstract}

\section{Introduction}
In the recent years machine learning (ML) algorithms made several breakthroughs in issuing accurate predictions. There is however a growing need to improve trustworthiness of these models. Providing accurate predictions is not enough in high-stake applications like healthcare where an agent (e.g. clinician) needs to interact with the model. In these applications the agent usually needs to understand how a particular prediction is issued. Especially, if the model prediction (say treatment plan) is different from what the clinician has in mind, explaining the model is vital. Complicated ML models like neural networks are essentially black-boxes to humans, and that is why interpretability methods are important and have gained significant attention in recent years~\cite{ribeiro2016lime,lundberg2017shap,guidotti2018lore,arnaldo2014multiple,zhang2018interpreting,alvarez2018towards,arrieta2020explainable,lou2013accurate,doshi2017towards}.
%There are different approaches for providing interpretation for ML models. One method is to provide local approximation of the model. Some of the most popular methods like LIME \cite{ribeiro2016lime}, SHAP \cite{lundberg2017shap}, and DeepLIFT \cite{arnaldo2014multiple} are this category. They explain the model by providing instance-wise feature importance. Another approach is global explanation. Some of the early works on interpretability which focus on approximating neural networks with a decision tree (which is considered interpretable) \cite{ifthen1995extracting,DT1996extracting} lies in this category. Using parzen window \cite{baehrens2010explain}... Other methods for explanation includes providing

There exist two key approaches to bring interpretability to machine learning models: (1) by designing inherently interpretable models \cite{Rudin_naturemi_2019,Chen_neurips_2019,Melis_neurips_2018}; or (2) by designing post-hoc methods to understand a pre-trained model \cite{Ribeiro_icmlws_2016,Lipton_icmlws_2016}. In this work, we focus on the second approach that includes methods to analyze a trained model locally and globally \cite{Montavon_dsp_2018}. The local interpretability methods focus on instance-wise explanations, which although useful, provide little understanding of a model's global behaviour \cite{ribeiro2016lime,lundberg2017shap}. Hence, researchers have proposed multiple techniques to interpret how a ML model behaves for a group of the instances. Some examples of global analysis methods include permutation feature importance \cite{Molnar_book_2019}, activation-maximization \cite{Erhan_tr_2009}, and learning globally surrogate models \cite{ifthen1995extracting,DT1996extracting}.

Our work relates to the last approach that aims to learn interpretable proxies by approximating the behaviour of black-box ML models for multiple instances. Some efforts in this category include methods to approximate neural networks with if-then rules~\cite{ifthen1995extracting} or decision trees~\cite{DT1996extracting} and the method to approximate matrix factorisation models using Bayesian networks and simple logic rules~\cite{Sanchez_aaai_2015}. Our work is mostly relevant to a different category of approaches for learning interpretable proxies that focuses on approximating black-box functions with symbolic metamodels.
%One way of providing interpretation which is mostly relevant to this work is to approximate the back-box function with a metamodel. 
A proper interpretable metamodel can enjoy benefits of different categories of interpretability methods. For example, a metamodel may provide insight into the interactions of different features and how they contribute in producing results. The metamodel can be locally approximated (e.g. using Taylor series) to generate instance-wise explanations. Moreover, it may be used for scientific discovery by revealing underlying laws governing the observed data \cite{schmidt2009distilling,wang2019symbolic,udrescu2020symbolic}.

Symbolic regression (SR) \cite{koza1994genetic}, has been the primary approach for finding approximate metamodels. In SR, there exist some fixed mathematical building blocks (e.g. summation operation), and the Genetic Programming (GP) algorithm searches over possible expressions that can be composed by combining the building blocks.
%In SR, there exist some fixed mathematical building blocks (e.g. summation operation), and the algorithm search over possible expressions that can be composed by combining these blocks. Genetic programming (GP) is the algorithm proposed to search over the possible configurations in SR. 
We will explain SR  \rev{in more details} in Section~\ref{sec:prelims} and compare it with our \rev{proposed} method in Section \ref{sec:related}. The major limitation of SR is that it uses a set of limited predefined building blocks and the search spaces grows when the number of building blocks increase. Two recent papers, which are the most relevant to our work \cite{alaa2019demystifying,crabbe2020learning}, address this issue by suggesting the use of a parametric trainable class of functions instead of fixed building blocks. In particular, they suggest using Meijer G-functions (we briefly introduce this class in Section~\ref{sec:prelims}). %They consider this class of function 
Note that these are univariate functions, in order to use them in multivariate settings, \cite{alaa2019demystifying} considers a heuristic approximation of Kolmogorov superposition theorem (KST) and \cite{crabbe2020learning} considers the projection pursuit method (in Section 4, we show that their method can be also considered as an approximation of KST). %Although the framework they use is general, both \rev{of these} works make some restricting assumptions that limit the \rev{usability and} coverage of their methods.
\rev{Both these works start from a general framework, however, they make some restricting assumptions that limit the usability and coverage of their methods.} For example, the simple function $x_1x_2$ (here $x_i$'s are features) cannot be represented with the method given in \cite{crabbe2020learning}. Similarly, the method in \cite{alaa2019demystifying} fails to represent  the product of three features $x_1x_2x_3$. Another limitation of the proposed approaches is that although most of familiar functions are indeed special cases of Meijer G-functions, for almost all parameters, Meijer G-functions do not have familiar closed form representation. Therefore, in practice, in the training of parameters it is very unlikely to obtain a set of parameters that are ``interpretable''.

%In this work, we address the above limitations by considering a more general approximation of KT and also by considering a simple class of functions to train. Our contribution is as follows, we represent expression of Kolmogorov theorem with a tree and approximate it by considering a particular set of sub-trees. We suggest using simple class of parameterized functions, each . We then suggest a version of GP appropriate for our setup to search over different trees and classes. Each step of evolution process of GP is augmented by using gradient descent to optimize the parameters of corresponding classes.
\textbf{High level idea and contribution:} In this work, we address the above challenges by proposing a new methodology to learn symbolic metamodels. Our approach is a generalization of \cite{alaa2019demystifying} and \cite{crabbe2020learning} as we consider a more general approximation of KST (see section \ref{sec:related}). We represent the KST expression using trees where edges represent simple parameterized functions (e.g., exponential function). We use gradient descent to train parameters of these functions and employ GP to search for the tree that most accurately approximates the black-box function. We demonstrate the efficacy of our proposed method through several experiments. The results suggest that our approach for estimating symbolic metamodels is comparatively more generic, accurate, and efficient than other symbolic metamodeling methods. In this work we are using our proposed method to provide interpretations, however, this method can be considered in general as a new GP method. Our method should be classified as a memetic algorithm where a population based method is paired with a refinement method (in our case gradient descent) \cite{chen2011multi}. To the best of our knowledge, this is the first method that uses gradient descent not only for training numerical constants but also for the training of building blocks, i.e., primitive functions.
%Discussing necessity of interpretability and its importance, a short introduction to different kinds of interpretations and some famous methods. Usefulness of metamodels as interpreter. Briefly explaining two previous models and their limitations. Highlighting contribution (GP with training instead of randomly choosing), it can be used for scientific discovery (SR has been already used). and organisation of the paper.

\section{Preliminaries}
\label{sec:prelims}
In this section, we present a brief overview of building blocks of our proposed method: genetic programming; and classes of trainable functions.\\
\textbf{Genetic Programming and symbolic regression:}
Genetic programming (GP) is an optimization method inspired by law of natural selection proposed by Koza in 1994 \cite{koza1994genetic}. It starts with a population of random programs for a particular task and then evolves the population in each iteration with operations inspired by natural genetic processes. The idea is that after enough iterations the population evolves and a fit program can be found in later generations. The two typical operations for evolving are crossover and mutation. In crossover, we choose the fittest programs (the fitness criterion is predefined for the task in hand) for reproduction of next generation (parents) and swap random parts of the selected pairs. In mutation operation, a random part of a program is substituted by some other randomly generated part of a program. One instance of using GP is for optimization in Symbolic Regression (SR), where the goal is to find a suitable mathematical expression to describe some observed data. In this setting, each program consists of primitive building blocks such as analytic functions, constants, and mathematical operations. The program is usually represented with a tree, where each node is representing one of the building blocks. We refer to \cite{orzechowski2018we,wang2019symbolic} for more details on SR. GP as a population base optimization method can be paired with other refinement methods. For example, here we are using both GP and GD in our model. This type of methods are called memetic algorithms. In particular, our method should be classified as a \textit{Lamarckian} memetic algorithm, where Lamarckian refers to the method of inheritance in GP search. we refer to \cite{emigdio2014evaluating,chen2011multi} for more details on taxonomy of GP methods. %The search over the resulting mathematical expressions for finding a suitable candidate is done via GP. %No particular model is provided as a starting point to the algorithm. Instead, initial expressions are formed by randomly combining mathematical building blocks such as mathematical operators, analytic functions, constants, and state variables. Instead, one provides a mathematical expression space containing candidate function building blocks, e.g. mathematical operators, state variables, constants, analytic functions, and then symbolic regression searches through the space spanned by these primitive building blocks to find the most appropriate solution.

% \textbf{Kolmogorov superposition theorem:}
% Kolmogorov superposition theorem \cite{kolmogorov1957} states that any multivariate continuous function with $d$ variables has a representation in terms of univariate functions as follows:
% \be g(\xb)=g(x_1,\cdots,x_d)=\sum_{i=1}^{2d+1}g^{out}_i\left(\sum_{j=1}^d g^{in}_{ij} (x_j)\right). \label{eq:Kol} \ee

\textbf{Class of trainable functions:}
\rev{In contrast with SR which uses fixed building blocks, our proposed approach (similar to \cite{alaa2019demystifying} and \cite{crabbe2020learning}) uses a class of trainable parameterized functions as building blocks.}
%Our proposed approach uses parameterized functions to represent the edges of random trees. \rev{In principle,} any class of functions that could be optimised using an optimisation algorithm (e.g., gradient descent) can be used as functions in our approach.
One such class of functions is called Meijer G-functions and have been used in two recent approaches to learn symbolic metamodels~\cite{meijer1946,meijer1936uber}. A Meijer-G function $G^{m,n}_{p, q}$ is defined as an integral along the path $\Lc$ in the complex plane.
\begin{align*}
    &G^{m,n}_{p, q}\left(^{a_1,\ldots,a_p}_{b_1,\ldots,b_q}\,\Big|\, x\right) =\\ &\frac{1}{2\pi i} \int_{\mathcal{L}} \frac{\prod^m_{j=1} \Gamma(b_j - s)\prod^n_{j=1} \Gamma(1 - a_j + s)}{\prod^q_{j=m+1} \Gamma(1 - b_j + s)\prod^p_{j=n+1} \Gamma(a_j + s)}\, x^s\, ds,
\end{align*}  
% \be \resizebox{.99\hsize}{!}{$G^{m,n}_{p, q}\left(^{a_1,\ldots,a_p}_{b_1,\ldots,b_q}\,\Big|\, x\right) = \frac{1}{2\pi i} \int_{\mathcal{L}} \frac{\prod^m_{j=1} \Gamma(b_j - s)\prod^n_{j=1} \Gamma(1 - a_j + s)}{\prod^q_{j=m+1} \Gamma(1 - b_j + s)\prod^p_{j=n+1} \Gamma(a_j + s)}\, x^s\, ds,$} \nonumber\ee
where $0\leq m\leq q$ and $0\leq n\leq p$ are all integers, and $a_i,b_j\in \mathbb{R}$ for $1\leq i\leq p$ and $1\leq j\leq q$. $\Lc$ is a path which separates poles of $\Gamma(1 - b_j + s)$ from poles of $\Gamma(a_j + s)$. 
By fixing $m,n,p,q$ we have a class of parameterized functions ($a_i$'s and $b_i$'s are parameters), which can be trained using gradient descent. We refer to \cite{beals2013meijer} for a more detailed definition of these functions. Meijer G-functions are rich set of functions that have most of the familiar functions which we think of as interpretable as special cases. For example, \be \resizebox{.99\hsize}{!}{$G^{0,1}_{3, 1}(^{2,2,2}_{\, \,\,\,\,1}\,|\,x) = x, \hspace{2mm} G^{1,0}_{0, 1}(\,^{-}_{\,0}\,|\,x) = e^{-x}, \hspace{2mm}  G^{1,2}_{2, 2}(^{1,1}_{1,0}\,|\,x) = \log(1+x).$} \nonumber\ee
However, when trained using gradient descent (GD), the final parameters for Meijer G-functions almost always will not have an interpretable closed form. This limits insight into  the functional form of the black-box model.
Hence, in this work, we propose using classes of simple, interpretable,  parameterized functions that can be efficiently optimized using GD. The class of functions can be chosen by a domain expert for each particular task. We will discuss the selection of primitive functions further in Appendix C. Specifically, here we demonstrate our approach using the following five parameterized functions. In Appendix C, we show that our presented results will not significantly change with using other set of primitive functions.
%Hence, in this work, we demonstrate our approach using the class of simple parameterized functions that can be efficiently optimised using GD. Specifically, our approach uses five parameterized functions (mentioned below) represented by edges in random trees.
\begin{align*}
&f_1(a, b, c, d|x)=ax^3+bx^2+cx+d,\,f_2(a,b|x)= ae^{-bx}\\
&f_3(a,b,c|x)=a \sin(bx+c), \, f_4(a,b,c|x)=a\log(bx+c), \\ 
&f_5(a,b,c,d|x)= ax/(bx^2+cx+d).   
\end{align*}

%The polynomial function can be thought of as Taylor approximation of degree three, and other four functions are added as we expecting . 

\textbf{Remark.} It is important to revisit that our proposed framework is generic and can accommodate any trainable class of functions, including Meijer G. %However, we leave the exploration of other classes of functions for future work.

%Another appeal of using Meijer G-functions is that we can use gradient descent to train its parameters. In this work similar to \cite{alaa2019demystifying} and \cite{crabbe2020learning} we consider $m,n,p$, and $q$ as hyperparameters and $a_i$'s and $b_i$'s as the parameters of the function. We will further discuss computing gradients with respect to $a_i$'s and $b_i$'s in the next section. We denote by $\Gc^{m,n}_{p, q}$ the set of all Meijer G-functions with a given $m,n,p$, and $q$. For a set of 4-tuples $\Hc$, $$\Gc_{\Hc}=\bigcup_{(m,n,p,q)\in \Hc}\Gc^{m,n}_{p, q}.$$ 
%In this work we will use a predefined $\Hc$ as hyperparameter space and search over $\Gc_{\Hc}$ to approximate the black box function. We use the same hyperparameter space that \cite{crabbe2020learning} introduces. That is, we choose 
%$$\Hc=\{(1,0,0,2),(0,1,3,1),(2,0,1,3),(2,1,2,3),(2,2,3,3)\}.$$ 
%In \cite[proposition 3.1]{crabbe2020learning} it is proved that $\Gc_{\Hc}$ includes all the functions with the following format:
%$$f(z)=\Phi(wz^q)z^t$$
%where $p,q,w\in \mathbb{R}$ and $\Phi$ can be any exponential, sinusoidal, identity, and their inverse functions. It can also be Bessel function or Euler's Gamma function. We refer to \cite{crabbe2020learning} for details about this hyperparameter space.
\section{Method}
Assume that a black box function $f:\Xc \to \mathbb{R}$ is trained on a dataset. Our goal is to find an interpretable function $g$ which approximates $f$. To this end, we restrict $g$ to belong to the class of functions $\Gc$ which are deemed to be interpretable. Therefore, we want to find the solution to the following optimization problem:
\be \label{eq:opt} \argmin_{g\in \Gc} \ell (f,g),\ee
where $\ell$ is our loss function of choice. In this work, we assume $\ell$ to be mean square loss
\be\ell (f,g)= \int_{\Xc} (g(x)-f(x))^2 dx. \label{eq:loss}\ee
%Using class of Meijer G-functions allow parameterizing of functions which can be optimized using gradient descent. 
In order to approximate multivariate function $f$, we deploy Kolmogorov superposition theorem \cite{kolmogorov1957} which states that any multivariate continuous function (with $d$ variables) has a representation in terms of univariate functions as follows:
\be g(\xb)=g(x_1,\cdots,x_d)=\sum_{i=1}^{2d+1}g^{out}_i\left(\sum_{j=1}^d g^{in}_{ij} (x_j)\right). \label{eq:Kol} \ee
In our setting, each of $g_{ij}^{in}$ and $g_i^{out}$ can be a function from $\Gc$. %Now given the wealth of functions that are included in the class of Meijer G-functions, and the fact that we can parameterize these functions and optimize one idea would be using these functions for implementing \eqref{eq:Kol}. 
However, fully implementing this equation (especially, using computationally expensive Meijer G-functions) is impractical even for moderate values of $d$. Therefore, an approximation is proposed in \cite{alaa2019demystifying} by considering a single outer function which is set to be identity and adding multiplication of all pairs of attributes to capture their correlation (we discuss this method in more details in Section \ref{sec:related}). In this work we propose another method for approximating Equation \eqref{eq:Kol}. 

%We suggest constructing a particular format of random trees which similar to classical symbolic regression are optimized using GP. Each edge in the tree will represent a Meijer G-function. In the following section we describe how to construct and optimize these trees.
\subsection{Approximating KST}
\begin{figure}[t]
%\begin{wrapfigure}{r}{0.5\textwidth}
\centering
\begin{tikzpicture}[scale=0.5,shorten >=1pt]
  \tikzstyle{vertex}=[circle,fill=black!25,minimum size=20pt,inner sep=2pt]
  \node[vertex] (G_1) at (0,0) {$x_1$};
  \node[vertex] (G_2) at (2,0)   {$x_2$};
  \node[vertex] (G_3) at (4,0)  {$x_3$};
  \node[vertex] (G_4) at (6,0)   {$x_4$};
  \node[vertex] (G_5) at (8,0)  {$x_5$};
  \node[vertex] (G_6) at (10,0) {$x_6$};
  \node[vertex] (G_7) at (12,0)  {$x_7$};
  \node[vertex] (G_8) at (3,2) {$h_1$};
  \node[vertex] (G_9) at (6,2)  {$h_2$};
  \node[vertex] (G_10) at (9,2) {$h_3$};
  \node[vertex] (G_11) at (6,4)  {$r$};
  \foreach \from/\to in {G_1/G_8,G_1/G_9,G_2/G_8,G_3/G_10,G_4/G_8,G_5/G_10,G_5/G_9,G_6/G_10,G_7/G_9,G_7/G_10,G_8/G_11,G_9/G_11,G_10/G_11}
  \draw (\from) -- (\to);
\end{tikzpicture}
\caption{A sample tree structure, each edge is representing a univariate function} \label{fig:structure}
\vspace{-10pt}
%\end{wrapfigure}
\end{figure}
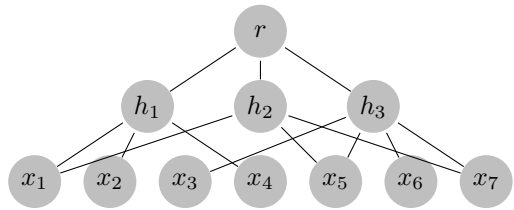

In our method, we approximate KST using trees with $L<2d+1$ middle nodes, where each of them is connected to only a subset of inputs. We denote the middle nodes with $h_i$, for $1\leq i \leq L$. \rev{Our approximation} can be represented \rev{via a} three layered tree (see Figure \ref{fig:structure}). There is a single root node at the top of the tree which is connected to $L$ middle nodes. Each middle node is connected to a subset of bottom layer nodes. The bottom layer of the tree has $d$ nodes corresponding to $d$ features. For simplicity, when it is not confusing, we call the node corresponds to $i$th feature by $x_i$. 

Note that each edge in the graph represents a univariate function. We denote the function corresponding to the edge between $h_i$ and the root with $g_{h_i}$ \rev{(these are the outer functions)}, and the function corresponding to an edge between $h_i$ and $x_j$ is denoted by $g_{ij}$ \rev{(inner functions)}. The argument of $g_{ij}$ is naturally the feature it is connected to, namely $x_j$, and the argument of $g_{h_i}$ is the summation of all incoming functions to $h_i$. That is, 
$\sum_{j\in \mathcal N(h_i)} g_{ij} (x_j),$
where $\mathcal{N}(h_i)$ denotes the neighbours of node $h_i$ \rev{in the graph}.
% There are $1\leq L$ nodes in the middle layer (denoted by $h_1,\cdots, h_L$) and 
%a single root node at the top layer. 
%Each of the edges are representing a Meijer G-function where the argument is the value of the bottom node (which can be $x_j$'s or $h_i$'s). 
Finally, for the root node we sum all the outputs of all $L$ middle layer functions. Therefore, each tree is representing a function from $\Xc$ to $\mathbb{R}$, which can be expressed as follows:
\be g(\xb)= \sum_{i=1}^L g_{h_i}\left(\sum_{j\in \mathcal N(h_i)} g_{ij} (x_j) \right).\label{eq:main}\ee
\subsection{Using GP for Training of Metamodels}
Now we want to solve the optimization problem in \eqref{eq:opt}, where $\Gc$ is the set of all functions that can be represented in form of Equation \eqref{eq:main}, where all $g_{ij}$ and $g_{h_i}$ are drawn from the class of \rev{primitive parameterized functions.} We propose solving this optimization problem by running a version of genetic programming algorithm. The tree representation of Equation \eqref{eq:main}, resembles the trees in symbolic regression that represents each program. Note that, unlike normal GP, here our constructed trees has a fixed structure of three layers, and also edges are representing functions. Hence we need to modify GP accordingly. In this section, we explain the details of our algorithm. 

%The idea is that, we can search over different trees and also different hyperparameters using a GP-like optimization algorithm. 
%There are $d$ nodes at the bottom layer, each corresponds to one of the attributes of the feature. 
%\be g(\xb)= \sum_{i=1}^L g_{h_i}\left(\sum_{j\in \mathcal N(h_i)} g_{ij} (x_j) \right).\label{eq:main}\ee
%Note that the full implementation of equation \eqref{eq:Kol} using Meijer G-functions is equivalent of having $L=2d+1$, and connecting all the nodes in bottom layer to middle layer nodes.

%For example, the standard form of Kolmogorov superposition theorem given in Eq 3 of the paper, is a two-layer tree where $l_1=2d$ and $l_2=1$. The heuristic method suggested in the paper is not in exactly in this format, however, if you also append pairs of inputs to input layer, then they have a single layer tree.
\subsubsection{Producing random trees}
In the first step, we produce $M$ random trees $T_1, \cdots,T_M$. 
% with the following structure. Inspired by Kolmogorov superposition theorem we will only focus on two-level trees. There is a single node in the highest level of the tree which is connected to all middle layer nodes. 
Each tree $T_i$ has $L_i$ middle nodes, where $L_i$ is an integer in $[l_1,l_2]$. $l_1$ and $l_2$ are important hyperparameters, determining number of middle nodes. %We denote these middle nodes by $h_1,h_2,\cdots,h_{L_i}$. The bottom layer of the tree are $d$ nodes corresponding to $d$ features, for simplicity when it is not confusing we call the node corresponds to $i$th feature $x_i$. 
For each of $L_i$ middle nodes, a random subset of bottom layers will be chosen to be connected to this node. At first instance, for all $1\leq u\leq L_i$ and $1\leq v \leq d$, we connect $h_u$ and $x_v$ with probability $0<p_0$. Then if there exist an $x_v$ which is not connected to any of the middle nodes. We choose $1\leq u\leq L_i$ uniformly at random and then connect $x_v$ and $h_u$ to ensure every $x_i$ is connected to at least one of the middle nodes. $p_0$ is the parameter that controls sparsity of the produced graphs, which is one of the main factors that determine the complexity of the training procedure. Each edge is representing a function from our class of primitive functions, thus we uniformly at random choose one of the function classes for each edge and also initialize its parameters with samples from normal distribution.
%Each edge is representing a Meijer G-function, we should hence assign a hyperparameter to all the edges. For each of them we uniformly at randomly choose one of the five hyperparameter setting in $\Hc$.

%(in expectation, degree of $h_1$ will be $p_0d$). Now for $h_2$, if $x_i$ was connected to $h_1$ then it will be connected to $h_2$ with probability $p_{-1}$ whereas if it was not connected the probability of an edge between the two is $p_1$. In general, probability of connection of $x_i$ to $h_j$ depends on the number of edges between $x_i$ and $h_1$ to $h_{j-1}$, if there are $e$ edges then it will be $p_{j-e-2}$. The set of $p_i$'s is predefined
%$$\mathcal{P}=\{p_{-L+1},\cdots, p_{-1},p_0,p_1,\cdots,p_{L-1}\}.$$
%We choose $p_{L-1}=1$, this guarantees that each attribute is at least connected to one of the middle nodes.

%It can be a number between $1$ to $d_{max}$, where $d_{max}$ is a hyperparameter. In layer $i$ of the tree there are up to $l_i^{max}$ nodes. Note that the last layer always has one node. Now for a node in $i$th layer we assign a random binary vector $(v_1,\cdots, v_{l_{i-1}})$ of length $l_{i-1}$ ($l_{i-1}$ is the number of nodes in the previous layer). Where $v_j=1$ shows that the node is connected to $j$th node of layer $(i-1)$.
\subsubsection{Training phases}
In the training phase, for each tree, we update the parameters of each edge using gradient descent. We choose a constant $k$ and apply $k$ gradient descent updates on the parameters of functions $g_{h_i}$ and $g_{ij}$. Let $g'_{h_i}(x) = \frac{d g_{h_i} (x)}{d x}$. For $a$ one of the parameters of $g_{ij}$ and $b$ a parameter of $g_{h_i}$, the gradient of $g$ with respect to $a$ and $b$ can be computed as follows \rev{(recall that $g$ is representing the metamodel)}:
\begin{align}
&\frac{\partial g(\xb)}{\partial a}=\frac{\partial g_{ij}(x_j)}{\partial a} \cdot g'_{h_i}\left(\sum_{k\in \Nc(h_i)} g_{ik}(x_k)\right),\\
&\frac{\partial g(\xb)}{\partial b}=\frac{\partial g_{h_i}}{\partial b}\left(\sum_{j\in \Nc(h_i)} g_{ij}(x_j)\right).
\end{align}  
In this work, we choose a fixed learning rate and leave the exploration of using more advanced optimization techniques for future work \rev{(this is compatible with \cite{alaa2019demystifying} and \cite{crabbe2020learning}, and allows us to have a fair comparison with these works)}.
% When $c$ is a parameter of $g_{h_i}$ we have:
% \be \frac{\partial g(\xb)}{\partial c}=\frac{\partial g_{h_i}}{\partial c}\left(\sum_{t\in \Nc(h_i)} g_{it}(x_t)\right). \ee
% For a Meijer G-function, derivative with respect to $x$ can be computed as follows \cite{beals2013meijer}:
% \be x^h\frac{d^h}{d x^h}G^{m,n}_{p, q}\left(^{a_1,\ldots,a_p}_{b_1,\ldots,b_q}\,\Big|\, x\right)= G^{m,n+1}_{p+1, q+1}\left(^{0,a_1,\ldots,a_p}_{b_1,\ldots,b_q,h}\,\Big|\, x\right).\ee
% Also, for partial derivation with respect to the parameters of the function we can use the following approximation \cite{beals2013meijer}:
% \begin{align}
% \frac{\partial}{\partial a_k}G^{m,n}_{p, q}\left(^{a_1,\ldots,a_p}_{b_1,\ldots,b_q}\,\Big|\, x\right)&\approx x^{a_k-1}G^{m,n+1}_{p+1, q}\left(^{-1,a_1-1,\ldots,a_p-1}_{\hspace{7mm} b_1,\ldots,b_q}\,\Big|\, x\right),\\    
% \frac{\partial}{\partial b_k}G^{m,n}_{p, q}\left(^{a_1,\ldots,a_p}_{b_1,\ldots,b_q}\,\Big|\, x\right)&\approx x^{1-b_k}G^{m,n}_{p, q+1}\left(^{\hspace{15mm} a_1,\ldots,a_p}_{b_1-1,\ldots,b_m-1,0,b_{m+1}-1,b_q-1}\,\Big|\, x\right).
% \end{align} 

\subsubsection{Evaluation fitness of metamodels} 
For evaluating fitness of the trained \rev{metamodels}, we uniformly at random sample $m$ points from $\Xc$ and query the output of black-box $f$ and metamodels $g_1, \cdots, g_M$ on these $m$ points and compute the mean square loss for the metamodels to approximate \eqref{eq:loss} (the output of $f$ is considered as the ground truth). If any of the $M$ models has a loss less than a predefined threshold we terminate the algorithm. Otherwise, we choose the $s$ fittest metamodels and discard the rest. These $s$ survived metamodels are the parents that will populate the next generation of trees in the evolution process for the next round of the algorithm. 

\textbf{Regularization:} We can modify the fitness criterion to favor simpler models. For encouraging sparsity of the tree, we can add a term  to the MSE error for penalizing trees that have more edges. Denoting total number of edges with $E$, we use this criterion for evaluation fitness of the trees ($\lambda$ is a hyperparameter):
\be \text{Fitness of a given tree} = \text{MSE} + \lambda E.\ee
%It should be highlighted that the regularization term will be only used for selecting the surviving trees and not for the gradient descend part. Hence, we can flexibly modify this further if needed. For example, if there is a preference towards a particular primitive function (e.g. because they are simpler) we can reflect that in the above criterion.
\subsubsection{Evolution phase}
In the evolution phase, we create the next generation of metamodels using survived trees. Similar to conventional GP algorithm, here we also define two operations to perform on each tree: Crossover and Mutation. For each of the $s$ chosen trees like $T$, we first pass on $T$ to the next generation, then we randomly choose $\frac{M}{s}-1$ times one of the two operations, perform it on $T$, and add the resulting tree to the cohort of the next generation trees. Thus, the total number of trees in the next cohort is also $M$. Here we define the two operations which preserve the three layer structure of the trees:
\begin{itemize}
\item In the crossover operation, \rev{for $T$,} we first randomly choose one of the nodes at the second layer of $T$. Then we uniformly at random choose one of the other $s-1$ trees, and then again uniformly at random choose one of its second layer nodes and replace that node alongside with all edges connected to that node with the chosen node in $T$. Notice that the edge connected to the root node will be also replaced. \rev{Moreover,} note that the new tree will inherit the functions corresponding to replaced edges and their parameters. 
\item %In the mutation operation, for a given tree we choose a subset of its edges connecting the first and second layer of the tree (here we choose only a single edge) and then either change the hyperparameters for that edge, remove it, or leave it unchanged. More precisely, first we choose uniformly at random an element in the set $\mathcal{H}\cup \emptyset$. When $\emptyset$ is the chosen element, we remove the edge, if the same hyperparameter is chosen we do not change anything, finally if a new hyperparameter is chosen we select that hyperparameter for that edge and reinitialize the parameters of the Meijer G-function corresponding to the edge.
%In mutation operation, we choose a subset of edges and apply one of the following modification. Either we remove the edge or we change the class of edges. This will allow us to explore 
In the mutation operation, one of these two actions will be applied on the tree: 1) changing the function class of an edge, 2) removing an edge between the middle and input layers. %3) creating a new edge between the middle and input layers. 
In each round of mutation, we apply $n_m$ times one of these two actions on the tree. %(here, we do not create a new edge to encourage sparsity). 
When we change the class of function for an edge, we also randomly reinitialize the parameters of the corresponding function. 
\end{itemize}
The above two operations allow us to explore different configurations of trees and classes. A pseudo code of the algorithm and a flowchart is presented in Appendix A. We call our proposed method symbolic metamodeling using primitive functions (SMPF).
% \begin{figure*}[t!]
%     \centering
%     \includegraphics[scale=0.4]{flowchart1.pdf}
%     \caption{Caption}
%     \label{fig:my_label}
% \end{figure*}

% \subsection{Hyperparameter space}
% In this work we use the same hyperparameter space that \cite{crabbe2020learning} uses. That is, we choose 
% $$\Hc=\{(1,0,0,2),(0,1,3,1),(2,0,1,3),(2,1,2,3),(2,2,3,3)\}.$$ 
% In \cite[proposition 3.1]{crabbe2020learning} it is proved that $\Gc_{\Hc}$ includes all the functions with the following format:
% $$f(z)=\Phi(wz^q)z^t$$
% where $p,q,w\in \mathbb{R}$ and $\Phi$ can be any exponential, sinusoidal, identity, and their inverse functions. It can also be Bessel function or Euler's Gamma function. We refer to \cite{crabbe2020learning} for details about this hyperparameter space.

% \subsection{Regularization}
% We can modify the fitness criterion to favor simpler models. Firstly, for encouraging simpler hyperparameter setting, we can penalize each function with the number of zeros and poles of Meijer G-function, i.e., $p+q$. Denote with $H$ summation of $p+q$ for all the edges of the tree over $E$, number of edges in the tree we add $\lambda_1 H$ to the fitness equation. Also, for encouraging sparsity of the tree we can add another term for penalizing using extra edges $\lambda_2 \frac{E}{d}$ ($d$ is used for normalizing number of edges).  

% \be \text{Fitness of the metamodel} = \text{MSE} + \lambda_1 H+ \lambda_2 \frac{E}{d}\ee
\subsection{Different Types of Interpretation Using SMPF}\label{sec:taylor}
\textbf{Instance-wise feature importance:} Similar to \cite{alaa2019demystifying} and \cite{crabbe2020learning} we can use the learned metamodel for estimating instance-wise feature importance. We can find the Taylor expansion of the metamodel around the data point of interest $\xb_0$ and analyse its coefficients.
\be \resizebox{.98\hsize}{!}{$g(\xb)=g(\xb_0)+\nabla g(\xb_0).(\xb-\xb_0)+(\xb-\xb_0).H_x(\xb).(\xb-\xb_0)+ \cdots,$}\ee
first order partial derivative with respect to $j$th feature can be computed using chain rule:
\be \frac{\partial g(\xb)}{\partial x_j}=\sum_{h_i\in \Nc(x_j)}g'_{h_i}\left(\sum_{j\in \Nc(h_i)} g_{ij}(x_j)\right)g'_{ij}(x_j). \ee
We will use this method in the instance-wise experiment. Importantly, we can also compute higher order coefficients for analyzing feature interactions.

\textbf{Mathematical expressions:} The final expression of the metamodel can provide insights into the functional form of the black-box function. For example, in the first experiment, we show that the metamodel correctly identifies that the black-box is an exponential function. Moreover, the inspection of mathematical expressions provides information about the interactions between the input features, and %Thus, knowledge gained by visualising and understanding mathematical expressions 
can potentially lead to understanding of previously unknown facts about the underlying mechanisms to domain experts.
An idea for exploring in future work is inspecting the final cohort of graphs. For example, if in the last iteration, the average degree of a node is large across different graphs, this can show the importance of the corresponding feature. Similarly, when a subset of features are connected to a middle node it can show the interaction of those features.
\section{Comparison with Related Works}\label{sec:related}
In the experiments section, %in Section~\ref{sec:exps}, 
we compare our approach with three symbolic metamodeling methods. This section briefly introduces these approaches , highlighting their strengths and weaknesses. A table comparing our method with a wider range of methods is provided in the supplementary material.

\textbf{Symbolic Metamodeling (SM) \cite{alaa2019demystifying}:}
SM proposes using Meijer G-functions for interpreting black box models. In the derivation of their method, they also start with KST \eqref{eq:Kol}, however, with a different approximation: they consider only one outer function ($g^{out}$) and set that function to be identity (the inner functions are all Meijer G). This does not allow the features to interact, in order to fix this problem, they add multiplication of all pairs $x_ix_j$ to the features. This setting has two main issues, firstly this method cannot capture interaction of more than two features and does not show other forms of interactions apart from multiplication. Secondly, this approach introduces many new features which makes it impractical when $d$ increases. There are ${d \choose 2}+d$ features in total and there is a Meijer G-function corresponding to each of them which makes using SM computationally costly.

\textbf{Symbolic Pursuit (SP) \cite{crabbe2020learning}:}
SP is a subsequent work to SM and is designed to overcome some of its flaws. In particular, SP is designed to use fewer Meijer G-functions. The method is based on the Projection Pursuit algorithm in statistics \cite{friedman1981projection}. In each step of the algorithm, a Meijer G-function will be fitted which minimizes the residual error between the metamodel and the black-box. The final metamodel will be the summation of all these Meijer G-functions. The input of each function is a linear combination of features. Thus, the final function will have the following formulation:
\be g(\xb)=\sum_{i=1}^L g_i\left(\sum_{j=1}^d c_{ij}x_j\right), \label{eq:SP}\ee
where $g_i$'s are Meijer G-functions. Importantly, the authors use a modified version of \eqref{eq:SP} where the arguments of Meijer G-functions are normalized such that they lie in the open interval of 0 to 1. Moreover, SP involves adding weights to the outer summation to allow mitigating the contributions of previously found functions, if needed.

Note that SP can be considered as one instance of our framework. The equation \eqref{eq:SP} is compatible with KST \eqref{eq:Kol} and can be represented similar to Figure 1. In essence, all inner function (edges between bottom and middle layers) are restricted to be linear, basically, they are coming from class of $f(x)= cx$. There are $L$ middle nodes, and outer functions are drawn from the class of Meijer G-functions. Also, in their setup $p_0=1$ ($p_0$ was the probability of connecting two nodes). 
\rev{A major problem with SP is its capability in representing non-linear correlations between the features.} For example, a simple function like $x_1x_2$ cannot be represented in SP formulation. Therefore, when using SP for explaining this function, in the best case scenario, by inspecting $c_{i1}$ and $c_{i2}$ we can understand that these two features are important but we cannot see how they interact. This can be potentially resolved in our framework by using a more general class of functions as inner functions. %\rev{Another issue, as we mentioned in Section 2, is that Meijer G-functions when trained using GD, almost always, will not have closed form representation.}
\begin{table*}[t!]
\centering
\resizebox{0.73\textwidth}{!}{%
%{\scriptsize
\begin{tabular}{cccccc} 
\toprule
& &$f(\xb) = e^{-3x_0+x_1}$ & $f(\xb) = \sin (x_0x_1)$ & $f(\xb) = \frac{x_0x_1}{(x_0^2+x_1)}$ &  $f(\xb) = \text{sinc}(x_0^2+x_1)$ \\
\midrule   
SMPF & $\ba \mbox{MSE}\\ R^2\ea$ & $\ba  \textbf{0.001}\pm \textbf{0.0002}\\ \textbf{0.996} \pm \textbf{0.002}\ea$ & $\ba 0.012\pm 0.002\\0.962 \pm 0.004\ea $&$\ba \textbf{0.002}\pm \textbf{0.0004}\\ \textbf{0.895} \pm \textbf{0.013}\ea$ & $\ba \textbf{0.004} \pm \textbf{0.0004} \\  \textbf{0.952}\pm \textbf{0.003} \ea$ \\\midrule
SM & $\ba \mbox{MSE}\\ R^2\ea$  & $\ba0.174\pm 0.031\\0.273\pm 0.019\ea$   & $\ba 0.126\pm 0.009\\ -2.039 \pm 0.442\ea$  & $\ba 0.108\pm 0.0104\\-5.461\pm 0.746\ea$ & $\ba 0.193 \pm 0.006 \\  -0.263\pm 0.094\ea$\\\midrule 
SP & $\ba \mbox{MSE}\\ R^2\ea$   & $\ba 0.009\pm 0.004\\ 0.958\pm 0.014\ea$   & $\ba 0.0008\pm 0.0001\\ 0.978\pm 0.003\ea$  & $\ba 0.002\pm 0.0003\\0.878\pm 0.021\ea $ & $\ba 0.009 \pm 0.002 \\ 0.937\pm 0.015 \ea$ \\\midrule
SP$^p$ & $\ba \mbox{MSE}\\ R^2\ea $ & $\ba 0.009\pm 0.001\\ 0.953\pm 0.014\ea$   & $\ba 0.024 \pm 0.001 \\ 0.348 \pm 0.082 \\ \ea$  & $\ba 0.011\pm 0.001 \\ 0.345 \pm 0.807  \ea $ &  $\ba 0.010\pm 0.001 \\ 0.932 \pm 0.013 \ea $ \\\midrule
SR & $\ba \mbox{MSE}\\ R^2\ea $  & $\ba0.078\pm 0.018\\0.658\pm 0.032\ea$   & $\ba \textbf{0.0004}\pm \textbf{0.0002}\\ \textbf{0.988}\pm \textbf{0.003}\ea$  & $\ba 0.012\pm 0.002\\ 0.256 \pm 0.144\ea$ & $\ba 0.016 \pm  0.003 \\ 0.886\pm 0.034 \ea$\\
\bottomrule 
\end{tabular}}
\caption{Approximating two-variable functions using SM, SP, SR and SMPF. \vspace{1mm}}
\label{exp1-table}
\vspace{-4pt}
\end{table*}

\textbf{Symbolic Regression:}
We briefly introduced SR in Section 2. SR searches over mathematical expressions that can be produced by combining a set of predetermined functions. %This is one of the main limitations of SR. The complexity of the search increases as the number of building blocks (e.g. predefined functions) increases.%Thanks to Meijer G-functions we are able to tune the parameters of these functions which can produce infinitely many functions. 
In each program, the leaf nodes are either features or numerical values, and other nodes are mathematical operations. One main difference between SR and our method (also SM and SP) is that unlike SR our methods are based on a representation derived from KST. Furthermore, we use parametric functions \rev{(and GD)} which cannot be accommodated in SR setting \rev{(note that GD has been suggested in SR but only for training of leafs, e.g. see \cite{topchy2001faster,kommenda2018local}).}
Importantly, SR has an advantage over SP and SM that the final result expression is guaranteed to be explainable, as it will be a combination of functions that we chose to include as the building blocks. However, when Meijer G-functions are used (in SM and SP), the resulting metamodel may not have a simple and explainable representation. This issue is resolved in our framework. There are several extensions on the original SR method, including methods that leverage deep learning techniques for searching the search space. These methods can be considered for future work to improve the GP in our method as well \cite{arnaldo2014multiple,rad2018gp,wang2019symbolic,orzechowski2018we,chen2015,udrescu2020ai,petersen2019deep,mundhenk2021symbolic}.
\section{Experiments}
\label{sec:exps}
We evaluate and compare our proposed method using three experiments. In the first experiment, we use our method to approximate four functions with simple expressions (similar to first experiment of \cite{alaa2019demystifying}). In the second experiment, we use our method for estimating instance-wise feature importance for three synthetic datasets (similar to \cite{alaa2019demystifying} and \cite{chen2018learning}). Finally, in the third experiment, we consider black-boxes trained on real data and approximate it using  the metamodel (similar to \cite{crabbe2020learning}).  %This section provides the details of the experiments. 
Some additional results and the hyperparameters are reported in Appendix E.
\begin{figure*}[t!]
    \centering
    \includegraphics[width=0.94\textwidth]{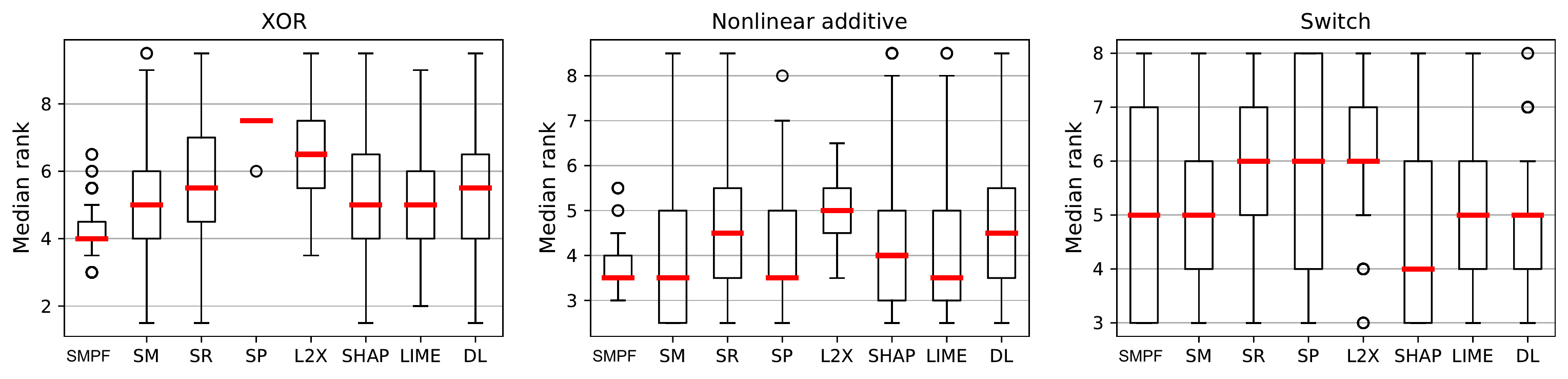}
\caption{Box-plot of feature importance for three datasets. The red lines show the median ranks under each algorithm. Lower median ranks imply better performance. DL refers to DeepLIFT.}
\label{fig:exp2}    
\end{figure*}
\subsection{Metamodels for Fixed Functions}
In this experiment, % similar to \cite{alaa2019demystifying}
\rev{we find metamodels for four synthetic functions with two variables.} We compare the performance of our method (SMPF) with symbolic metamodeling (SM), symbolic pursuit (SP), polynomial approximation of SP (SP$^p$), and symbolic regression \cite{orzechowski2018we} (similar to \cite{alaa2019demystifying} we use gplearn library \cite{stephens2015gplearn} for implementation of SR). We compare methods in terms of mean squared error (MSE) and $R^2$ score. %Specifically, we consider four synthetic functions: an exponential $e^{-3x_0+x_1}$, a sinusoid $\sin(x_0\cdot x_1)$, a rational $\frac{x_0x_1}{x_0^2+x_1}$, and a sinc function $\text{sinc}(x_0^2+x_1)$. 
Generally, our algorithm achieves a better accuracy as compared to other methods (we have the best score for three of the functions). The results are reported in Table 1.
Furthermore, SMPF was able to correctly identify the functional form. For the first experiment, the final expression of the metamodel is as follows (we rounded up coefficients here):
\begin{align*}
g(\xb) = 0.854\exp\Big(&-2.438\sin(1.371x_0 - 0.0318) +\\ &\frac{0.684x_1}{0.016x_1^2+0.204x_1+0.426}\Big).
\end{align*}
This shows an important advantage of our method in comparison to other methods.
For example, the expression found by SP algorithm has the following form ($P1$ here is a linear combination of the two inputs):
\begin{equation*} g(x) = 0.98\, G^{2,1}_{2, 3}\left(^{\,\,\,\, 0.24,-0.06}_{0.16,-0.47, 0.43}\,|\,1.0 [ReLU(P1)] \right).\end{equation*}
Note that it was not possible to find a closed form expression for this function. 
Also, for the second function, $\sin$ is correctly chosen as the outer function in SMPF. See Appendix E, where we provide results for synthetic functions with more variables.

\begin{table*}[t!]
\centering
%\resizebox{\columnwidth}{!}{%
{\scriptsize
\begin{tabular}{cccccc} 
\toprule
& Method &\multicolumn{2}{c}{MLP}& \multicolumn{2}{c}{SVM} \\ 
& & MSE & $R^2$ & MSE & $R^2$ \\ \midrule
Black Box && $  0.689 \pm 0.224  $&$  0.703\pm 0.019 $ &$ 0.448 \pm 0.241  $ & $ 0.781\pm 0.061  $\\\midrule
Method v.s. Black Box &$\ba \mbox{SMPF} \\ \mbox{SP} \ea$ & $\ba 0.007 \pm 0.003 \\ 0.008 \pm 0.011 \ea$ &$\ba 0.993\pm 0.003 \\ 0.978 \pm 0.016 \ea$& $\ba 0.029 \pm 0.013 \\ 0.014 \pm 0.015 \ea$ &$\ba 0.967\pm 0.120 \\ 0.974 \pm 0.078 \ea$ \\\midrule
Method &$\ba \mbox{SMPF} \\ \mbox{SP} \ea$ &$\ba 0.674 \pm 0.211 \\ 0.682 \pm 0.225 \ea$&$\ba 0.709\pm0.015 \\ 0.697 \pm 0.027 \ea$ & $\ba 0.344 \pm 0.163 \\ 0.471 \pm 0.253 \ea$&$\ba 0.829\pm0.037 \\ 0.780 \pm 0.048 \ea$ \\
\bottomrule 
\end{tabular}}%}
\caption{Interpreting black-boxes trained on real data using SMPF compared with SP\vspace{1mm}}
\label{exp3-table}
\vspace{-4pt}
\end{table*}
\subsection{Instance-wise Feature Selection}
In this experiment, we evaluate the performance of our method for estimating the feature importance by repeating the second experiment of \cite{alaa2019demystifying}. Three synthetic datasets are used: XOR,  Nonlinear additive features, and Feature switching. All three datasets have 10 features, in XOR, only the first two features contribute in producing the output. In Nonlinear additive features and switch datasets, the first four features and first five features are important, respectively.\footnote{See Appendix B of \cite{alaa2019demystifying} for more details.} First, we train a 2-layer neural network $f(\xb)$ with 200 hidden neurons for estimating the label of each data point. Then, we run our algorithm to find function $g(\xb)$ to estimate function $f(\xb)$. We consider the coefficient of each feature in the Taylor series of $g(\xb)$ as a metric for its importance. The larger the coefficient, the more important it will be. We rank the features based on their importance. We consider 1000 data points, repeat the process for each data point and find the median feature importance ranking. The median value of relevant features determines the accuracy of the algorithm; the smaller median rank implies a better accuracy. Figure \ref{fig:exp2} compares our algorithm with Symbolic Metamodeling (SM) \cite{alaa2019demystifying}, Symbolic Pursuit (SP) \cite{crabbe2020learning}, Symbolic Regression \cite{orzechowski2018we}, DeepLIFT \cite{shrikumar2017learning}, SHAP \cite{lundberg2017shap}, LIME \cite{ribeiro2016lime}, and L2X \cite{chen2018learning}. SMPF performs competitively comparing with other algorithms. For XOR dataset we have the best median rank, and we are among the best for nonlinear additive dataset. On Switch dataset, SMPF performs similar to other global methods, i.e., SM, SP, and SR which are our direct competitors. SHAP is the only algorithm that has a better performance on this dataset. 

\subsection{Black-box Approximation}
In this experiment, we evaluate performance of our model on interpreting a black-box trained on real data, replicating the second experiment of \cite{crabbe2020learning}. A Multilayer Perceptron (MLP), and Support Vector Machine (SVM) are trained as two black boxes using UCI dataset Yacht \cite{Dua:2019} (additional results are reported in Appendix E). In order to have the same setting as SP, we train the MLP and SVM models using the scikit-learn library \cite{buitinck2013api} with the default parameters. %We randomly split each dataset into a training dataset and a test dataset. 
We randomly use 80\% of the data points for the training of the black box model as well as SMPF model, and the remaining 20\% is used to evaluate the performance of the model. This procedure is repeated five times to report the averages and standard deviations. We report the mean squared error (MSE) and $R^2$ score of the MLP and SVM against the true labels, MSE and $R^2$ of the metamodel against the black-box models, and the MSE and $R^2$ of the metamodel against the true labels (see Table \ref{exp3-table}). We observe that both SP and SMPF have very good performance in approximating the black-box. Interestingly, SMPF outperforms the black-box on the test set for both models which may indicate that the black-box overfits the dataset, but SMPF does not, as it uses simple functions.

\section{Discussion}
\textbf{Complexity:} In terms of run-time, for the last experiment, the training of SP for the MLP black-box takes 215 minutes, while the training of our algorithm takes 45 minutes (both performed on a personal computer). The reason that SP is more computationally expensive is that SP has to evaluate Meijer G-functions in each iteration of their optimization process. Evaluating a Meijer G-function is very expensive and takes about 1 to 4 seconds depending on the hyperparameters (i.e., $m,n,p,q$). This observation implies that SMPF has lower computational complexity which allows us to handle more variables and also enables the possibility of using more complex trees, as we suggest later in the future work. However, this should be highlighted that our method (similar to other symbolic methods) is not appropriate for high dimensional data like images.

\textbf{Limitations:} Even though we showed the performance of our model through extensive numerical experiments, our method lacks theoretical guarantees (theoretical analysis is particularly challenging because of the use of GP). Another limitation (also inherited from GP) is that there are several hyperparameters in our model to specify structure of the tree. As discussed, symbolic metamodels cannot handle high dimension inputs. Finally, the richness of functions we can create is limited, this can be compensated using more complex classes of functions or more complex tree structures.

\textbf{Direct training vs  using black-box:}
A natural question is why not directly use the training data to train the metamodel (without using the black-box)? There are two reasons for why we have considered the black-box for training. One is from the user point of view, we may have been given a task of interpreting a black-box, i.e., the user’s question may be why this particular method is working, and not necessarily looking for another interpretable method. Secondly, and more importantly, we may not have access to the dataset for various reasons including privacy concerns. In this method we only need querying the black-box method and we can use random inputs (as many of them as we want). Directly using the dataset in all symbolic metamodeling methods (e.g. SR, SM, and SP) is also possible and can be relevant in many scenarios, e.g., discovering the underlying governing rules of a dataset \cite{udrescu2020ai,sahoo2018learning,makke2021symbolic}.
%One run of SP algorithm on personal computer took 
%In comparison with other interpretation methods note that estimating metamodel is a one time operation, and in the test time we simply use Taylor series at that instance to compute feature importance.

\textbf{Conclusion and future work:} 
\rev{%Comparing with SM and SP which use Meijer G-functions (a richer class of functions) with a more simplistic approximation of KST. However, here we are proposing using simpler more interpretable functions instead of Meijer G-functions, and we compensated the loss of generality caused by using simpler functions by deploying a much more general approximation of KST.
We proposed a new generic framework for symbolic metamodeling based on the Kolmogorov superposition theorem. We  suggested using simple parameterized functions to get a closed-form and interpretable expression for the metamodel. The use of simple functions may seem restrictive when compared with SM and SP which use Meijer G-functions (a richer class of functions). However, this is compensated in our framework with a better approximation of KST. We used genetic programming to search over different possible trees and also possible classes of functions. 
%There are several directions for the expansion of this work. We can consider more complex tree structures. For example, we can have trees with four layers instead of three, which allows us to construct more complex expressions. Another direction is considering other functions in our structure, e.g., using Meijer-G functions in our setup. 
%Another direction for future work is to improve the training phase of the trees. In this work, the optimization problem is non-convex, and gradient descent may not be able to find the global optimal point when we train a tree. This issue can be addressed by imposing convex relaxation or using more sophisticated non-convex optimization methods for gradient descent (in this work, we used a fixed learning rate for gradient descent). 
There are several directions for the expansion of this work: 1) we can consider a more complex tree structure. For example, we can have trees with four layers instead of three, which allows us to construct more complex expressions (see Appendix D). 2) Other primitive functions can be used in our setup, e.g., Meijer G-functions. 
3) The optimization in the training phase can be improved. The problem is non-convex, and gradient descent may not be able to find the global optimal point. This issue can be addressed by imposing convex relaxation or using more sophisticated non-convex optimization methods.} %for gradient descent (in this work, we used a fixed learning rate for gradient descent). 

\textbf{Acknowledgment: } This work is partially supported by the NSF under grants IIS-2301599 and ECCS-2301601.

\bibliography{ref}

\end{document}

\maketitle
% \section{Ethics and Reproducibility Statements}
% \textbf{Ethics Statement:} This should be highlighted that we do not claim that the functional form found via our method is necessarily the correct function. This is the case even if the metamodel has a very good performance in approximating the black box.

% \textbf{Reproducibility Statement:} We have included the code for all three experiments in the supplementary material. We also, included a detailed description in a read me file, explaining how to run the code. Details of the experiments and hyperparameters are reported in Appendix \ref{app:exp}.
\appendix
%\section{Appendix}
\section{Pseudo code and flowchart of our algorithm}\label{app:pseudo}
In this section, we summarize our algorithm in the following pseudo code. For the notational convenience, we denote $g_{T_i}$ as the metamodel corresponding to tree $T_i$, $E_i$ as the number of edges in tree $T_i$, $\mbox{Thr}$ as a fixed threshold, $\mbox{lr}$ as the learning rate, $\mbox{Max\_Itr}$ as the maximum number of iterations, and $\mbox{MSE}_i = \frac{1}{n}\sum_{j=1}^n (g_{T_i}(x_i)-f(x_i))^2$, where $\{x_1,\ldots,x_n\}$ is the training dataset. %The next pseudo code summarizes the SMPF algorithm. 
\begin{algorithm}[h]
\SetAlgoLined
\KwInput{Black-box: $f(x)$; Training set: $\{x_1,\ldots,x_n\}$; $\newline$ Hyperparamters: $M, l_1, l_2, s, p_0, k, \mbox{Thr}$, $\mbox{lr}$, $\lambda$,  $p_{\text{cross}}$, $p_{\text{del}}$, $\mbox{Max\_itr}$. } 
\KwOutput{Metamodel $g(x)$ approximating function $f(x)$ }
Generates $L_i, i=1,\ldots,M$, where $L_i$ is a random integer between $l_1$ and $l_2$.\; 
Generates $M$ random trees $T_1,\ldots, T_M$ using parameters $L_i$ and $p_0$ (details provided in Section 3.2.1)\;
$\mbox{Itr}\leftarrow 0$\;
\While{$\mbox{Itr}\leq \mbox{Max\_itr} ~~\text{   and   }~~ \min_i \mbox{MSE}_i + \lambda E_{i} \geq \mbox{Thr}$   }{
\For{i = 1, 2,\ldots, M}
{
Update the parameters of the functions corresponds to the edges of tree $T_i$ using $k$ gradient descent updates and learning rate $\mbox{lr}$\;
%Calculate $\nabla l(f,g_{T_i})$ with respect to $a_1,\ldots, a_p, b_1,\ldots, b_q$, and run gradient descent with $k$ iterations to find the optimal values for  $a_1,\ldots, a_p, b_1,\ldots, b_q$ \;
Calculate the mean squared error ($\mbox{MSE}_i$) corresponds to tree $T_i$\;
$\mbox{Fitness}_i \leftarrow \mbox{MSE}_i + \lambda \cdot E_i$\;
}
Select $s$ trees with the smallest $ \mbox{Fitness}_i $ and re-index them by $T_1,\ldots,T_s$ \;
\For{$i = 0, 1,\ldots, s-1$}
{ 
    \For{$j = 1, 2,\ldots, M/s - 1 $}
    { 
    Generate random number $u$ uniformly in $[0,1]$\;
      \eIf{$u<p_{\text{cross}}$}{
    Generate tree $T_{i \cdot (M/s - 1) + j + s}$ by crossover operation on tree $T_{i+1}$\;
   }
   {
   Generate tree $T_{i \cdot (M/s - 1) + j + s}$ by mutation operation on tree $T_{i+1}$. In this step, an edge is removed with probability $p_{\text{del}}$\;
  }
  }
  }
  $\mbox{Itr}\leftarrow \mbox{Itr}+1$\;
}
$i^* \leftarrow \arg\min_{i} \mbox{MSE}_i + \lambda E_{i}$\;
$g(x)\leftarrow g_{T_{i^*}}$\;
\caption{Pseudo code for SMPF}\label{Alg1}
\end{algorithm}

The flowchart of our algorithm is given in the following figure:
\newpage
\begin{figure}[t!]
    \centering
    \includegraphics[width=0.8\textwidth]{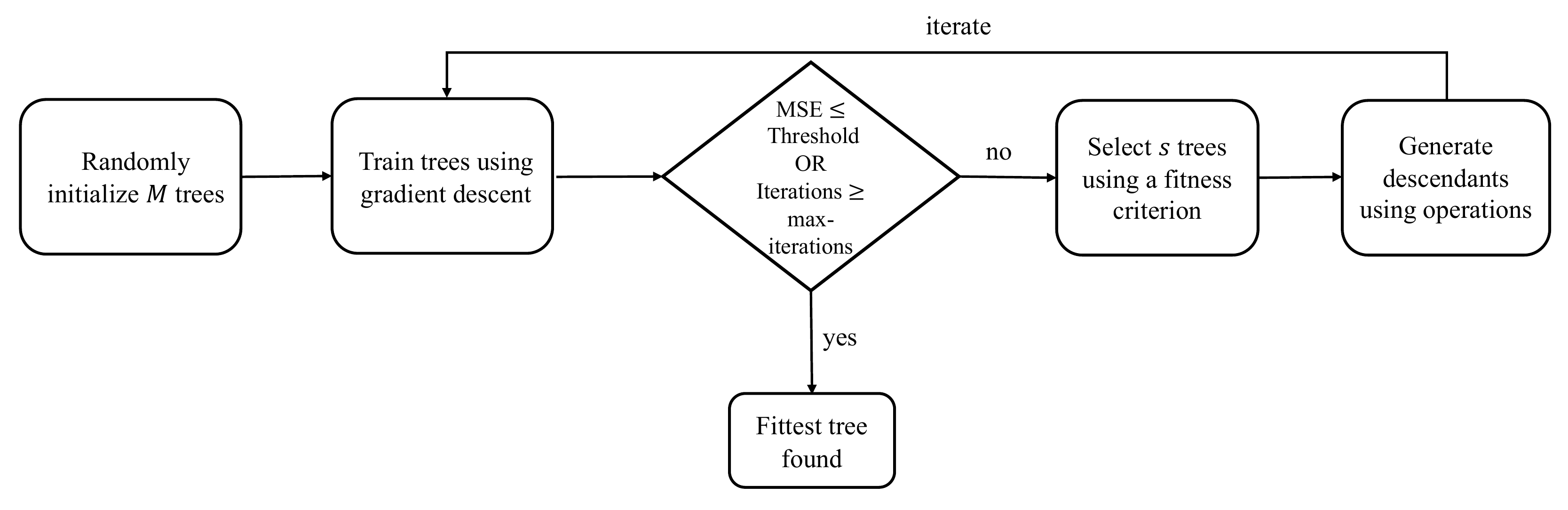}
    \caption{Flowchart of our proposed memetic algorithm}
    \label{fig:flowchart}
\end{figure}

%\newpage
\section{Related work table}\label{app:related}
In this section we provide a comparison table for several important interpretation methods. We adopted columns of this table from \cite{crabbe2020learning}[Table 3]. As it can be seen all of the approaches can provide feature importance, some of them are also capable of providing feature interactions. Most of the methods are indeed model independent. The subset of the methods that are global is determined. Finally, the interpretable expression here means whether the method is able to produce a closed-form expressions (this is only applicable for symbolic approaches). %GA$^2$M \cite{lou2013accurate}, IG \cite{Sundarrajan_icml_2017}, and LRP \cite{Bach_plos_2015} refer to generalised additive models with interactions, integrated gradients and layer-wise relevance propagation methods, respectively. 

\begin{table}[h!]
\centering
%\resizebox{\columnwidth}{!}{%
\caption{Comparison of interpretability methods\vspace{1mm}}
{\scriptsize
\begin{tabular}{|c|c|c|c|c|c|} 
\hline
Algorithm &Feature importance&Feature Interaction& Model independent & Global & Interpretable expression\\ \hline
LIME    \cite{ribeiro2016lime} & \checkmark& &\checkmark & &\\\hline
SHAP \cite{lundberg2017shap}    & \checkmark& \checkmark&\checkmark & &\\\hline
DeepLIFT \cite{shrikumar2017learning}& \checkmark& & & &\\\hline
GA$^2$M  \cite{lou2013accurate}   & \checkmark& \checkmark&\checkmark & &\\\hline
IG \cite{Sundarrajan_icml_2017} & \checkmark& & & &\\\hline
LRP \cite{Bach_plos_2015} & \checkmark& & \checkmark& &\\\hline
L2X  \cite{chen2018learning}    & \checkmark& &\checkmark & &\\\hline
SM  \cite{alaa2019demystifying}     & \checkmark& \checkmark&\checkmark &\checkmark &\\\hline
SP   \cite{crabbe2020learning}    & \checkmark& \checkmark&\checkmark & \checkmark&\\\hline
SP$^\text{p}$ \cite{crabbe2020learning}& \checkmark& \checkmark&\checkmark &\checkmark &\checkmark\\\hline
SR  \cite{koza1994genetic}     & \checkmark& \checkmark&\checkmark &\checkmark  &\checkmark\\\hline
SMPF     & \checkmark& \checkmark &\checkmark & \checkmark&\checkmark\\\hline
\end{tabular}}%}
\label{table-comparison}
\vspace{-4pt}
\end{table}
\section{Primitive functions}\label{app:prim}
In this section we discuss the selection of primitive functions. An advantage of our framework is its generality and the fact that it can work with any set of parameterized functions as long as its parameters can be trained using gradient descent. The primitive functions, in principle, can be chosen by the domain expert for the particular task in hand. For example, in the main text we considered five familiar functions as our primitive functions. The polynomial function there, can be thought of as Taylor approximation (hence it is reasonable to include it). The other four functions can be thought of as domain expert suggestion. 

To show that our results is not particularly dependent on the choice of primitive functions. Here, we repeat our experiments for two more set of primitive functions. In the first set we are using only three of the five functions: polynomial, exponential, and sinusoidal. In the second set, we keep the polynomial function and replace the other two functions with new parameterized functions.
\begin{align*}
\text{Set 1:} \quad f_1(a,b|x)= ae^{-bx}, \quad &f_2(a,b,c|x)=a \sin(bx+c),\nonumber \\
 f_3(a, b, c, d|x)&=ax^3+bx^2+cx+d.\nonumber
\end{align*}
\begin{align*}\text{Set 2:} \quad f_1(a, b, c, d|x)=ax^3+bx^2+cx+d&, \quad f_2(a,b,c|x)=a \arctan{(bx+c)}, \nonumber \\
f_3(a, b, c, d|x)&=\frac{ax+b}{cx+d} \nonumber.
\end{align*}
In Table \ref{app:exp1-table}, we report result of the first experiment for these two sets of primitive functions. It is interesting to note that removing $\sin$ function from Set 2 resulted in an inferior performance for $\sin$ and $\text{sinc}$ functions. Also, inclusion of the function with fraction form ($f_3(a, b, c, d|x)=\frac{ax+b}{cx+d}$) in Set 2, improved the performance of this set on the third function.
\begin{table}[h!]
\centering
\caption{Approximating two-variable synthetic functions \vspace{1mm}}
{\scriptsize
\begin{tabular}{cccccc} 
\toprule
& &$f(\xb) = e^{-3x_0+x_1}$ & $f(\xb) = \sin (x_0x_1)$ & $f(\xb) = \frac{x_0x_1}{(x_0^2+x_1)}$ &  $f(\xb) = \text{sinc}(x_0^2+x_1)$ \\
\midrule   
Set 1 & $\ba \mbox{MSE}\\ R^2\ea$ & $\ba  0.0004 \pm 0.00008\\ 0.998 \pm 0.0002\ea$ & $\ba 0.0023 \pm 0.0004\\0.947 \pm 0.0123\ea $&$\ba 0.002 \pm 0.0006 \\ 0.762 \pm 0.042\ea$ & $\ba0.001 \pm 0.0002 \\  0.991 \pm 0.002 \ea$ \\\midrule
Set 2 & $\ba \mbox{MSE}\\ R^2\ea$  & $\ba 0.004\pm 0.001\\0.980\pm 0.003\ea$   & $\ba 0.007\pm 0.0007\\ 0.840 \pm 0.024\ea$  & $\ba 0.004\pm 0.0004\\0.790\pm 0.028\ea$ & $\ba 0.018 \pm 0.005 \\  0.876\pm 0.043\ea$\\
\bottomrule 
\end{tabular}}
\label{app:exp1-table}
\vspace{-4pt}
\end{table}

We have also repeated the experiment 3. The result for Yacht dataset is presented in Table \ref{app:exp3-yacht}. Both sets of functions perform reasonably. Set 1 functions approximates the black-box better, however, Set 2 functions have a better performance on the test dataset. The performance of original set of 5 functions and SP method is reported in Table \ref{exp3-table}. 

We also report here the results for Energy dataset (another UCI dataset used in \cite{crabbe2020learning}). It can be seen that there was not a significant difference in most of the results. The performance of original set of 5 functions and SP method is reported below in Table \ref{exp3-table-energy}. %only the third synthetic function has a worse performance which is not surprising as we removed the fraction function. %Note that there is only one middle node in the first experiment, so there is no way to produce surrogates for the fraction function.
\begin{table}[h!]
\centering
%\resizebox{\columnwidth}{!}{%
\caption{The result of two sets of primitive functions on Yacht dataset \vspace{1mm}}
{\scriptsize
\begin{tabular}{cccccc} 
\toprule
& Method &\multicolumn{2}{c}{MLP}& \multicolumn{2}{c}{SVM} \\ 
& & MSE & $R^2$ & MSE & $R^2$ \\ \midrule
Black Box && $  0.625 \pm 0.153  $&$  0.725 \pm 0.018 $ &$ 0.674 \pm 0.356  $ & $ 0.754 \pm 0.059 $\\\midrule
Method v.s. Black Box &$\ba \mbox{Set 1} \\ \mbox{Set 2} \ea$ & $\ba 0.009 \pm 0.005 \\ 0.014 \pm 0.008 \ea$ &$\ba 0.992 \pm 0.005 \\ 0.989 \pm 0.005 \ea$& $\ba 0.033 \pm 0.017 \\ 0.045 \pm 0.018 \ea$ &$\ba 0.967 \pm 0.012 \\ 0.951 \pm 0.010 \ea$ \\\midrule
Method &$\ba \mbox{Set 1} \\ \mbox{Set 2} \ea$ &$\ba 0.648 \pm 0.170 \\ 0.627 \pm 0.073 \ea$&$\ba 0.720 \pm 0.030 \\ 0.728 \pm 0.015 \ea$ & $\ba 0.557 \pm 0.257  \\ 0.483 \pm 0.253 \ea$&$\ba 0.794 \pm 0.037 \\ 0.798 \pm 0.054 \ea$ \\
\bottomrule 
\end{tabular}}%}
\label{app:exp3-yacht}
\vspace{-4pt}
\end{table}

\begin{table}[h!]
\centering
\caption{The result of two sets of primitive functions on Energy dataset\vspace{1mm}}
{\scriptsize
\begin{tabular}{cccccc} 
\toprule
& Method &\multicolumn{2}{c}{MLP}& \multicolumn{2}{c}{SVM} \\ 
& & MSE & $R^2$ & MSE & $R^2$ \\ \midrule
Black Box && $  0.016 \pm 0.002  $&$  0.919 \pm 0.013  $ &$ 0.011 \pm 0.001  $ & $ 0.945 \pm 0.003 $\\\midrule
Method v.s. Black Box &$\ba \mbox{Set 1} \\ \mbox{Set 2} \ea$ & $\ba 0.003 \pm 0.001 \\ 0.004 \pm 0.002 \ea$ &$\ba 0.982 \pm 0.009 \\ 0.973 \pm 0.013 \ea$& $\ba 0.007 \pm 0.001 \\ 0.007 \pm 0.001 \ea$ &$\ba 0.961 \pm 0.005 \\ 0.963 \pm 0.009 \ea$ \\\midrule
Method &$\ba \mbox{Set 1} \\ \mbox{Set 2} \ea$ &$\ba 0.019 \pm 0.003 \\ 0.020 \pm 0.002 \ea$&$\ba 0.903 \pm 0.020 \\ 0.898 \pm 0.008 \ea$ & $\ba 0.017 \pm 0.002 \\ 0.016 \pm 0.003 \ea$&$\ba 0.914 \pm 0.009 \\ 0.917\pm0.012 \ea$ \\
\bottomrule 
\end{tabular}}%}
\label{exp3-app-energy}
\vspace{-4pt}
\end{table}

\subsection{Future work: ideas for primitive functions selection:} 
Here we introduce two ideas for the selection of primitive functions. We leave exploration of usability of these methods for future work. 

One possibility is generalizing the SP formulation \cite{crabbe2020learning}. As we explained in Section \ref{sec:related}, SP method can be regarded as an special case of our framework where the outer functions are chosen from Meijer G-functions and the inner functions are coming from the class of $f(x)=cx$. An immediate generalization is to consider more general inner functions. For instance, we can replace linear functions with polynomial functions of higher degree, or consider a set of classes to choose from. 

Another possibility is including polynomials of variable degree among primitive functions. For example, for each function, we can randomly determine the degree by sampling from a decaying distribution. This allows us to occasionally include a polynomial of higher degree (i.e., a Taylor approximation with more accuracy). In order to avoid overfitting we can penalize the fitness criterion of each tree with the degree of the polynomial that it uses. 

%Another strategy is to consider primitive functions as hyperparameters of the function and run few different set of functions.

\section{Extension to more complex trees}\label{app:tree}
Given a set of primitive functions, we can extend the set of possible metamodels that can be produced by these primitive functions by adding new layers to the tree structure. This extension enables us to create more complex metamodels which may be required for some black-boxes. For instance, in Figure \ref{app:fig}, we show a sample tree with four layers. Similarly, if needed, we can add new additional layers (resembling hidden layers) to increase the capacity of the model.

Each hidden layer will have $L_i$ nodes. We produce edges randomly, similar to three-layered trees, we can independently connect two nodes with probability $p_0$ (or we can have different connection probabilities for different layers). Also, we can similarly define two evolution operations. In the mutation operation, we randomly select some of the edges and either change the corresponding function or delete that edge. For crossover operation, we randomly choose one of the middle nodes, it can be from any of the hidden layers, and change all of its edges with another middle node (in the same layer) of another surviving tree.

The resulting equation of the metamodel is given in \eqref{eq:app:layer}:
\begin{figure}[h!]
%\begin{wrapfigure}{r}{0.5\textwidth}
\centering
\begin{tikzpicture}[scale=0.7,shorten >=1pt]
  \tikzstyle{vertex}=[circle,fill=black!25,minimum size=20pt,inner sep=2pt]
  \node[vertex] (G_1) at (0,0) {$x_1$};
  \node[vertex] (G_2) at (2,0)   {$x_2$};
  \node[vertex] (G_3) at (4,0)  {$x_3$};
  \node[vertex] (G_4) at (6,0)   {$x_4$};
  \node[vertex] (G_5) at (8,0)  {$x_5$};
  \node[vertex] (G_6) at (10,0) {$x_6$};
  \node[vertex] (G_7) at (12,0)  {$x_7$};
  \node[vertex] (G_8) at (2,2) {$h^1_1$};
  \node[vertex] (G_9) at (5,2)  {$h^1_2$};
  \node[vertex] (G_10) at (8,2) {$h^1_3$};
  \node[vertex] (G_11) at (11,2) {$h^1_4$};
  \node[vertex] (G_12) at (4,4)  {$h^2_1$};
  \node[vertex] (G_13) at (8,4)  {$h^2_2$};
  \node[vertex] (G_14) at (6,6)  {$r$};
  \foreach \from/\to in {G_1/G_8,G_1/G_9,G_2/G_8,G_3/G_10,G_4/G_8,G_4/G_11,G_7/G_11,G_5/G_10,G_5/G_9,G_6/G_10,G_7/G_9,G_7/G_10,G_8/G_12,G_9/G_12,G_9/G_13,G_10/G_13,G_11/G_12,G_11/G_13,G_12/G_14,G_13/G_14}
  \draw (\from) -- (\to);
\end{tikzpicture}
\caption{The structure of a sample four layer tree. Adding additional layers can allow us to create more complex metamodels.} \label{app:fig}
\vspace{-10pt}
%\end{wrapfigure}
\end{figure}
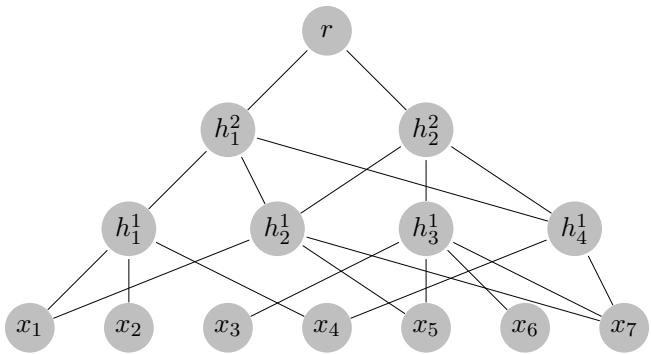
\be g(\xb)= \sum_{i=k}^{L_2} g_{h^2_k}\left(\sum_{i\in \mathcal N(h^2_k)}g_{h^1_i} \left(\sum_{j\in \mathcal N(h^1_i)} g_{ij} (x_j) \right)\right).\label{eq:app:layer}\ee
\section{Experiments}\label{app:exp}
In this section we provide the details of our three experiments as well as some additional results.
\subsection{Experiment 1}
We used the following hyperparameters: $M=k=20$, recall that $M$ is the population of each generation, and $k$ is the number of gradient updates in each iteration. We choose learning rate to be $0.1$ and the maximum number of iterations to be $30$. Also, $s=4$, so in each iteration we choose $4$ best trees to populate next generation. Here we have only one input, i.e. $l_1 =l_2 =1$. Hence, there are exactly three edges in the graph (also we do not remove any edges in the mutation operation), and the GP algorithm in this experiment only searches over various possible ways for assigning classes to edges.

Here we also provide results of an additional experiments similar to first experiment with functions with three variables. We modified the first function in experiment 1 as follows:
\begin{equation*}
f(\xb)=e^{-3x_0 + x_1 + x_2^2}.
\end{equation*}
For this function (with same hyperparameters as before), our model was able to find a metamodel with $R^2 = 0.958 \pm 0.010$. Also, the functional form is correctly chosen:
\begin{equation*}
g(\xb) = 2.82\exp(-1.10x_0^3 - 0.48x_0^2 - 1.96x_0 - 1.46\exp(-1.04x_1) + 1.27\sin(0.97x_2 + 0.05)).
\end{equation*}
Similarly, we added a variable to the second function, and used our model for approximating it.
\begin{equation*}
f(\xb)= \sin(x_0x_1 + x_2).
\end{equation*}
Again our model correctly identified the functional form with $R^2= 0.902\pm 0.018$:
\begin{equation*}
g(\xb)=1.00 \sin(0.18x_0^3 + 0.31x_0^2 + 0.015x_0 + 0.28x_2^3 + 0.41x_2^2 + 0.47x_2 + 0.50\sin(0.97x_1 + 0.01) - 0.07).
\end{equation*} 

\textbf{Extension of Experiment 1:} Here we consider synthetic functions with more than two variables. We consider functions with 4, 6, and 8 variables, and study how the performance of each model changes when we increase the dimension. Here in Table 7, for brevity we only report results of SP and SR, and exclude SM as it did not perform well even for two variables (as shown in Table 1). We have considered three following sets of functions, similar to Section 5.1, we consider an exponential function, a sin function and a ratio of two functions:

$$e_4(\xb)=e^{-2 x_1 + x_2 + 3x_3 - 3 x_4} \quad e_6(\xb)=e_4(\xb)e^{-2x_5+x_6} \quad e_8(\xb)=e_6(\xb) e^{-x_7x_8}$$
$$s_4(\xb)=\sin(x_1x_2 + x_3 + 2x_4) \quad s_6(\xb)=\sin(x_1x_2 + x_3 + 2x_4 + 3x_5 - x_6)$$
$$s_8(\xb)=\sin(x_1x_2 + x_3 + 2x_4 + 3x_5 - x_6+\sqrt{x_7}x_8)\quad r_4(\xb)= \frac{x_1x_2}{(x_1^2 + x_2 + x_3^2 + 2x_4)} $$
$$r_6(\xb)= \frac{x_1x_2}{(x_1^2 + x_2 + x_3^2 + 2x_4+x_5^3+x_6)} \quad r_8(\xb)= \frac{x_1x_2+\sqrt{x_7}x_8}{(x_1^2 + x_2 + x_3^2 + 2x_4+x_5^3+x_6)}$$

It can be seen that SMPF has the most robust results, and adding variables does not significantly deteriorate the performance. We have noted that SP's performance is particularly dependent on the random seed (RS). For example, with the default RS=50, the R$^2$ score for $e_4(\xb)$ was $-0.008$, in Table 7 we reported the performance for RS=10 (R$^2=0.649$). It can also be seen that while the performance of the model is fine for $s_6(\xb)$ and $s_8(\xb)$, it is unexpectedly low for $s_4(\xb)$ (we believe this is a result of using Meijer G-functions which might be very dependent on the initialization). Also, we have observed that the difference between the training error and test error for SR is significantly large which suggest overfitting.

\begin{table}[t!]
\centering
\caption{Approximating multivariable functions using SM, SP, SR and SMPF. \vspace{1mm}}
{\scriptsize
\begin{tabular}{cccccc} 
\toprule
& &SMPF & SP & SP$^p$ &  SR \\
\midrule   
$e_4(\xb)$ & $\ba \mbox{MSE}\\ R^2\ea$ & $\ba  0.119\pm 0.082\\ 0.979\pm 0.007\ea$ & $\ba 1.992 \pm 0.430\\0.649 \pm 0.041\ea $&$\ba 2.112\pm 0.653\\ 0.623 \pm 0.039\ea$ & $\ba 2.632 \pm 0.492 \\  0.412\pm 0.054 \ea$ \\\midrule
$e_6(\xb)$ & $\ba \mbox{MSE}\\ R^2\ea$  & $\ba 0.174 \pm 0.031\\0.953\pm 0.029\ea$   & $\ba 0.198\pm 0.044\\ 0.432 \pm 0.052\ea$  & $\ba 0.201\pm 0.041\\ 0.427\pm 0.050\ea$ & $\ba 0.209 \pm 0.088 \\  0.401\pm 0.094\ea$\\\midrule 
$e_8(\xb)$ & $\ba \mbox{MSE}\\ R^2\ea$   & $\ba 0.009\pm 0.004\\ 0.958\pm 0.014\ea$   & $\ba 0.028\pm 0.011\\ 0.498\pm 0.033\ea$  & $\ba 0.029\pm 0.011\\0.496\pm 0.033\ea $ & $\ba 0.033 \pm 0.032 \\ 0.248\pm 0.065 \ea$ \\\midrule
$s_4(\xb)$ & $\ba \mbox{MSE}\\ R^2\ea $ & $\ba 0.003\pm 0.001\\ 0.950\pm 0.008\ea$   & $\ba 0.068 \pm 0.020 \\ -0.097 \pm 0.047 \\ \ea$  & $\ba 0.071\pm 0.019 \\ 0.101 \pm 0.045  \ea $ &  $\ba 0.010\pm 0.008 \\ 0.726 \pm 0.010 \ea $ \\\midrule
$s_6(\xb)$ & $\ba \mbox{MSE}\\ R^2\ea $  & $\ba 0.016\pm 0.001\\0.964\pm 0.002\ea$   & $\ba 0.022\pm 0.009\\ 0.948\pm 0.022\ea$  & $\ba 0.026\pm 0.009\\ 0.923 \pm 0.021\ea$ & $\ba 0.036 \pm  0.009 \\ 0.728\pm 0.022 \ea$\\\midrule
$s_8(\xb)$ & $\ba \mbox{MSE}\\ R^2\ea $ & $\ba 0.015\pm 0.002\\ 0.976\pm 0.006\ea$   & $\ba 0.018 \pm 0.002 \\ 0.955 \pm 0.020 \\ \ea$  & $\ba 0.021\pm 0.002 \\ 0.901 \pm 0.024  \ea $ &  $\ba 0.033\pm 0.007 \\ 0.744 \pm 0.033 \ea $ \\\midrule
$r_4(\xb)$ & $\ba \mbox{MSE}\\ R^2\ea $ & $\ba 0.002\pm 0.001\\ 0.793\pm 0.024\ea$   & $\ba 0.001 \pm 0.0001 \\ 0.883 \pm 0.025 \\ \ea$  & $\ba 0.001\pm 0.0001 \\ 0.861 \pm 0.026  \ea $ &  $\ba 0.010\pm 0.002 \\ 0.118 \pm 0.049 \ea $ \\\midrule
$r_6(\xb)$ & $\ba \mbox{MSE}\\ R^2\ea $ & $\ba 0.001\pm 0.001\\ 0.723\pm 0.038\ea$   & $\ba 0.001 \pm 0.0001 \\ 0.811 \pm 0.022 \\ \ea$  & $\ba 0.001\pm 0.0002 \\ 0.808 \pm 0.025  \ea $ &  $\ba 0.017\pm 0.002 \\ -0.013 \pm 0.051 \ea $ \\\midrule
$r_8(\xb)$ & $\ba \mbox{MSE}\\ R^2\ea $ & $\ba 0.004\pm 0.003\\ 0.729\pm 0.079\ea$   & $\ba 0.004 \pm 0.002 \\ 0.779 \pm 0.052 \\ \ea$  & $\ba 0.005\pm 0.002 \\ 0.771 \pm 0.055  \ea $ &  $\ba 0.014\pm 0.001 \\ 0.002 \pm 0.037 \ea $ \\
\bottomrule 
\end{tabular}}
\label{exp1:ext-table}
\vspace{-4pt}
\end{table}

\subsection{Instance-wise feature selection}
For this experiment we choose the same set of hyperparameters for all three datasets. We had $M=20$, $s=2$, and $k=10$. Number of iterations were $10$, and $lr=0.01$. We had $p_0=0.7$ (probability of connecting a middle node to a feature node) and $p_{\text{cross}} = 0.7$ (the probability of choosing crossover operation). Thus, for each survived tree, we apply crossover with probability 0.7 and mutation with probability 0.3. For each mutation operation, with probability $0.5$ we deleted a random edge from middle to bottom layer, and with probability of $0.5$ we changed hyperparameters of one of the $E$ edges. We chose, $l_1=l_2=2$, therefore, all trees had exactly $2$ middle nodes.
% \begin{figure}
% \begin{center}
% \includegraphics[width=\textwidth]{exp2_fig.pdf}
% \caption{Box-plot of feature importance for three datasets. The red lines show the median ranks under each algorithm. Lower median ranks imply better performance. DL refers to DeepLIFT.}
% \label{fig:exp2}    
% \end{center}
% \end{figure}
\subsection{Experiment 3}
The hyperparameters for this experiment are as follows. Similar to other experiments we initialized with $M=20$ trees. We had $s=4$ survived trees and number of evolution iterations was $20$. With $k=20$ gradient updates, and $lr=0.05$. In this experiment we had $p_0=0.6$. Parameters regarding operations were similar to experiment 2.

Additionally, here we are providing results for Energy dataset (another UCI dataset). %We used similar hyperparameters, the only difference is that here $M=28$.
Here we also included the result for SR method (SM is not included as its performance is not meaningfully close to these methods). It can be seen that SMPF is performing competitively with SP. In particular, when tested on the real dataset. It should be mentioned that SMPF and SR has the advantage that unlike SP they can produce closed form expression.

\begin{table}[h!]
\centering
%\resizebox{\columnwidth}{!}{%
\caption{Interpreting black-boxes trained on Energy dataset\vspace{1mm}}
{\scriptsize
\begin{tabular}{cccccc} 
\toprule
& Method &\multicolumn{2}{c}{MLP}& \multicolumn{2}{c}{SVM} \\ 
& & MSE & $R^2$ & MSE & $R^2$ \\ \midrule
Black Box && $  0.015 \pm 0.002  $&$  0.924\pm 0.008 $ &$ 0.014 \pm 0.001  $ & $ 0.927\pm  0.003 $\\\midrule
Method v.s. Black Box &$\ba \mbox{SMPF} \\ \mbox{SP} \\ \mbox{SR} \ea$ & $\ba 0.005 \pm 0.002 \\ 0.001 \pm 0.001 \\ 0.012 \pm 0.012 \ea$ &$\ba 0.970\pm 0.011 \\ 0.996 \pm 0.001 \\ 0.933 \pm 0.067 \ea$& $\ba 0.007 \pm 0.004 \\ 0.004 \pm 0.001 \\ 0.011 \pm 0.001 \ea$ &$\ba 0.962\pm 0.023 \\ 0.978 \pm 0.001  \\ 0.935 \pm 0.003 \ea$ \\\midrule
Method &$\ba \mbox{SMPF} \\ \mbox{SP} \\ \mbox{SR} \ea$ &$\ba 0.019 \pm 0.003 \\ 0.020 \pm 0.001 \\ 0.026 \pm 0.016 \ea$&$\ba 0.903\pm0.010 \\ 0.901 \pm 0.006 \\ 0.865 \pm 0.067 \ea$ & $\ba 0.018 \pm 0.002 \\ 0.017 \pm 0.001 \\ 0.019 \pm 0.002 \ea$&$\ba 0.908\pm0.014 \\ 0.914\pm0.001 \\ 0.905\pm0.005 \ea$ \\
\bottomrule 
\end{tabular}}%}
\label{exp3-table-energy}
\vspace{-4pt}
\end{table}
\bibliographystyle{ieeetr}
\bibliography{ref}